\documentclass[journal]{IEEEtran}
\usepackage{graphicx}
\usepackage{color}
\usepackage{lineno,hyperref}
\usepackage{amsmath}
\usepackage{multirow}
\usepackage{algorithm}
\usepackage{algorithmicx}
\usepackage{acronym}
\usepackage{algpseudocode}
\usepackage{colortbl}

\usepackage{times}
\usepackage{epsfig}
\usepackage{graphicx}
\usepackage{amsmath}
\usepackage{amssymb}
\usepackage{makecell}
\usepackage{arydshln}
\usepackage[numbers,sort&compress]{natbib} 

\graphicspath{{Img/}}

\begin{document}

\title{HIPA: Hierarchical Patch Transformer for \\Single Image Super Resolution}

\author{Qing~Cai,~\IEEEmembership{Member,~IEEE}
	    Yiming~Qian,
	    Jinxing~Li,
	    Jun~Lyu,
        Yee-Hong~Yang,~\IEEEmembership{Senior Member,~IEEE}
        Feng~Wu,~\IEEEmembership{Fellow,~IEEE}
        and~David Zhang,~\IEEEmembership{Life Fellow,~IEEE}

\thanks{This work was supported in part by the National Science Foundation of China under Grant 62102338, Grant 61906162, and Grant 62172347; in part by the Natural Science Foundation of Shandong Province under Grant ZR2020QF031; in part by the Qingdao Postdoctoral Innovation Project under Grant QDBSH20230101001; in part by the CUHK(SZ)- Linklogis Joint Laboratory of Computer Vision and Artificial Intelligence; in part by the Shenzhen Institute of Artificial Intelligence and Robotics for Society; in part by the Shenzhen Research Institute of Big Data; and in part by the Natural Sciences and Engineering Research Council of Canada and the University of Alberta.(\textit{Corresponding author: David~Zhang, Jun Lyu})}
\thanks{Qing~Cai is with the Faculty of Information Science and Engineering, Ocean University of China, Qingdao, Shandong, 266100, China.}
\thanks{Yiming~Qian is with the Department of Computer Science, University of Manitoba, Winnipeg, Manitoba, R3T 2N2, Canada.}
\thanks{Jinxing~Li is with the School of Computer Science and Technology, Harbin Institute of Technology, Shenzhen, Guangdong, 518055, China.}
\thanks{Jun Lyu is with the School of Nursing, The Hong Kong Polytechnic University, Hong Kong(e-mail:ljdream0710@pku.edu.cn).}
\thanks{Yee-Hong~Yang is with the Department of Computing Science, University of Alberta, Edmonton, Alberta T6G 2E9, Canada.}
\thanks{Feng~Wu is with the School of Information Science and Technology, University of Science and Technology of China, Hefei, Anhui, 230026, China.}
\thanks{David Zhang is with the School of Data Science, The Chinese University of Hong Kong, Shenzhen, Guangdong 518172, China, also with the Shenzhen Institute of Artificial Intelligence and Robotics for Society, Shenzhen, Guangdong 518000, China, and also with the CUHK(SZ)- Linklogis Joint Laboratory of Computer Vision and Artificial Intelligence, Shenzhen, Guangdong 518172, China(e-mail: davidzhang@cuhk.edu.cn).}}

\markboth{IEEE TRANSACTIONS ON IMAGE PROCESSING}%
{Shell \MakeLowercase{\textit{et al.}}: Bare Demo of IEEEtran.cls for IEEE Journals}

\maketitle
\begin{abstract}
Transformer-based architectures start to emerge in single image super resolution (SISR) and have achieved promising performance. However, most existing vision Transformer-based SISR methods still have two shortcomings: (1) they divide images into the same number of patches with a \emph{fixed} size, which may not be optimal for restoring patches with different levels of texture richness; and (2) their position encodings treat all input tokens equally and hence, neglect the dependencies among them. This paper presents a HIPA, which stands for a novel Transformer architecture that progressively recovers the high resolution image using a hierarchical patch partition. Specifically, we build a cascaded model that processes an input image in multiple stages, where we start with tokens with small patch sizes and gradually merge them to form the full resolution. Such a hierarchical patch mechanism not only explicitly enables feature aggregation at multiple resolutions but also adaptively learns patch-aware features for different image regions, e.g., using a smaller patch for areas with fine details and a larger patch for textureless regions. Meanwhile, a new attention-based position encoding scheme for Transformer is proposed to let the network focus on which tokens should be paid more attention by assigning different weights to different tokens, which is the first time to our best knowledge. Furthermore, we also propose a multi-receptive field attention module to enlarge the convolution receptive field from different branches. The experimental results on several public datasets demonstrate the superior performance of the proposed HIPA over previous methods quantitatively and qualitatively. We will share our code and models when the paper is accepted. 
\end{abstract}

\IEEEpeerreviewmaketitle

\section{Introduction}
\label{sec_Int}
\begin{figure}
	\centering
	\includegraphics[width=0.44\textwidth]{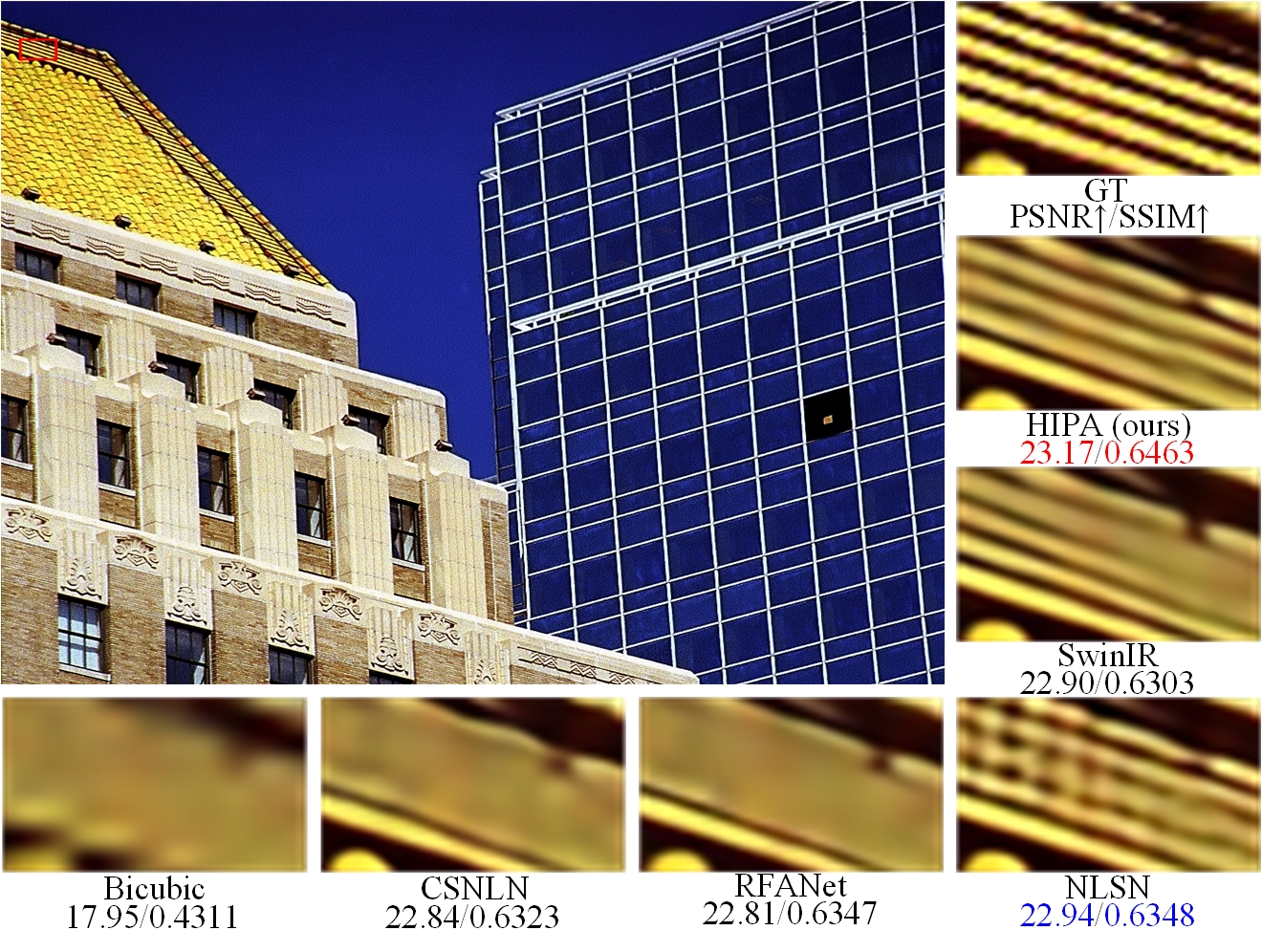}
	\caption{Visual comparisons of $\times$4 SISR on ``img\_063" from Urban 100. It can be seen that our method obtains better visual quality and recovers more textures and details compared with that of other state-of-the-art methods. The colors \textcolor{red} {red} and \textcolor{blue} {blue} represent the best and the second best methods.}
	\vspace{-4 mm}
	\label{fig_com}
\end{figure}

Single Image Super-Resolution (SISR), aiming to recover a high-resolution (HR) image from its corresponding degraded low-resolution (LR) version, plays an important and fundamental role in computer vision and image processing due to its wide range of real-world applications, such as medical imaging~\cite{li2021volumenet}, surveillance~\cite{wen2022video} and remote sensing~\cite{arefin2020multi}, amongst others. SISR is a very challenging and ill-posed problem because there is no unique solution for any given LR input~\cite{zhang2020deep,cai2022tdpn}.

Deep convolutional neural networks (CNNs) have achieved remarkable success in SISR and various architectures have been presented so far, for example, residual learning~\cite{anwar2019real,nie2020dual,wang2021adversarial}, dense connections~\cite{zhang2018residual,song2020efficient}, UNet-like architectures with skip connections~\cite{hu2019runet,prajapati2021channel}, dilated convolutions~\cite{yang2017deep,zhang2017learning}, generative models~\cite{ledig2017photo,wang2018esrgan,prajapati2021direct,wenlong2021ranksrgan} and other kinds of CNNs~\cite{guo2020closed,lu2021masa,guo2022data}. However, the convolution in CNN uses a sliding window to extract local features and hence, is weak in capturing long-range or non-local dependencies, which are important for SISR. In particular, for some regions with fine textures, faithful reconstruction depends not only on local relationships but also on long-range dependencies~\cite{zhang2021context,li2022transformer}. To alleviate this issue, many attention mechanisms have been proposed and introduced into SISR, such as global attention mechanism~\cite{zhang2018image,zamir2020cycleisp,niu2020single,liu2020residual,dai2019second} and non-local attention mechanism~\cite{zhang2019residual,mei2020image,mei2021image,xia2022efficient}. As shown in Fig.~\ref{fig_com}, although some state-of-the-art (SOTA) methods such as NLSN~\cite{mei2021image} could recover some amounts of high-frequency details, the reconstructed slanted line structures exhibit fuzzy and blurry boundaries, which are faithfully restored using our hierarchical patch Transformer.

Inspired by the significant success of Transformer in natural language processing~\cite{vaswani2017attention} for its advantages in modeling long-range context, vision Transformer is also introduced into the field of SISR~\cite{yang2020learning,chen2021pre,cao2021video,liang2021swinir,wang2022uformer} and has obtained superior results than many SOTA CNN-based methods due to the multi-head self-attention mechanism that is capable of modeling long-distance dependencies~\cite{chu2021twins}. Very recently, hybrid architectures combining CNN and Transformer start to emerge in the community~\cite{liang2021swinir} to fully utilize the advantage of CNN in extracting local features and the advantage of Transformer in establishing long-range dependencies. Although existing Transformer-based SISR models have achieved superior results, the recovered results still exhibit blurry boundaries as shown in the result of SwinIR~\cite{liang2021swinir} in Fig.~\ref{fig_com}. \textbf{The main reasons may lie in two shortcomings of existing vision Transformer-based SISR methods.}  First, almost all of them partition all input images into the same number of fixed-size patches, which may not be ideal considering different images on image regions have their own characteristics~\cite{wang2021not}. Second, the position encoding of most vision Transformer-based SISR methods treats all input tokens equally. However, the low-resolution input tokens contain abundant information for SISR, which are treated equally across tokens and hence, the representation ability of Transformer is limited.

In order to compensate the above two shortcomings, in this paper, we propose a \textbf{Hi}erarchical \textbf{Pa}tch (HIPA) Transformer by partitioning an input image into a hierarchy of patches with different sizes. In particular, a multi-stage architecture is first developed by alternately stacking CNN and Transformer to boost their benefits in feature extraction. Then, to achieve different size patch input for the Transformer and to let the Transformer establish global dependencies from different numbers of tokens, the LR image is first partitioned into a hierarchy of subblocks, which are used as inputs to the Transformer by starting from the small-size blocks and gradually merging them in the next stage. In addition, we design a novel attention-based position encoding scheme for the Transformer based on dilated channel attention to model the position information with a continuous dynamical model. Besides, a multi-receptive field attention module is proposed based on dilated convolution with different dilation factors to enlarge the convolution receptive field from different branches. As shown in Fig.~\ref{fig_com}, our HIPA obtains better visual quality compared with that of other state-of-the-art SISR methods.

Briefly, the contributions of this paper mainly include:
\begin{itemize}
	\item A novel hierarchical patch Transformer has been designed to achieve multi-size patches for Transformers. This approach is more effective than treating all samples with the same number of fixed-size patches because the hierarchical patch Transformer allows patches with different texture richness to adopt different sizes, rather a single size patch; 
	\item A new attention-based position encoding scheme is proposed for Transformer that allows the network to focus on which tokens should be paid more attention, which is the first time to our best knowledge;
	\item A multi-receptive field dilated attention module is designed to enlarge the convolution receptive field from different branches, which achieves relatively smaller increase of the computational complexity compared to the one by increasing the depth and the filter size of a CNN to enlarge the receptive field.
\end{itemize}

The rest of the paper is organized as follows: Section~\ref{sec_RW} briefly overviews related works. Section~\ref{sec_HIPA} presents the proposed HIPA Transformer and discusses its advantages and differences with existing methods. Section~\ref{sec_Exp} presents the experimental results and analysis of the proposed method by comparing it with state-of-the-art models. Finally, the paper concludes in Section~\ref{sec_Con}.

\section{Related Work}
\label{sec_RW}

\noindent \textbf{CNN-based Models:} The SRCNN model proposed by Dong \textit{et al.}~\cite{dong2014learning} is a pioneering work to apply CNN to single image super-resolution, which has achieved superior performance than traditional methods~\cite{zhang2006edge,zhang2012single,chen2015weighted} by using only a three-layer CNN to represent the mapping between LF and HR images. Based on the SRCNN, many deeper and wider CNN based SISR models have been proposed to achieve better restoration performance. However, blindly increasing the depth of a network does not necessarily improve the performance but may introduce many new issues for training, for example, the vanishing or exploding gradient~\cite{lim2017enhanced}. Later, residual learning is introduced into SISR to ease the training difficulty of deeper networks. For example, by introducing residual learning into a deeper network, Kim \textit{et al.} can stack more convolutional layers and propose the VDSR~\cite{kim2016accurate}. However, all of the above models need to first pre-process the LR input to obtain the desired image size using interpolation, which is not only time consuming but also often introduces noise and blurriness in the input image. To address the above issues, Dong \textit{et al.}~\cite{dong2016accelerating} introduce a deconvolution layer as the last layer and achieve end-to-end training for SISR. Such a deconvolution layer is then substituted by a more efficient sub-pixel convolution layer~\cite{shi2016real} proposed by Shi \textit{et al.}, which is also adopted by our method similar to the EDSR~\cite{lim2017enhanced} and the RCAN~\cite{zhang2018image}. However, all of these models treat the LR features equally across channels, which inevitably limits the restoration capability of CNNs. Even worse, the convolution kernel usually has a limited receptive field and cannot sufficiently extract long-range or non-local features. As a result, for some regions with fine details, these methods yield poor performance. 

\noindent \textbf{Attention-based Models:} To address the above issues, attention mechanism~\cite{zhang2018image,dai2019second,mei2020image,niu2020single} is introduced into SISR to guide the deep neural network to selectively pay more attention on features where there is more information. For example, by integrating channel attention and residual blocks, Zhang \textit{et al.} propose the RCAN~\cite{zhang2018image}, which markedly improves the representational performance of the CNN. Dai \textit{et al.} propose the SAN~\cite{dai2019second} using a novel trainable second-order channel attention. However, the channel attention treats different convolution layers independently and neglects the correlation among them. To alleviate this issue, Niu~\textit{et al.} propose the HAN~\cite{niu2020single} by integrating a layer attention module and a channel-spatial attention module into the residual blocks. More recently, non-local attention modules~\cite{zhang2019residual,mei2020image,mei2021image,xia2022efficient} are proposed to address the inherent issue of CNNs in establishing long range or non-local dependencies among exacted features. For example, Zhang~\textit{et al.}~\cite{zhang2019residual}, propose the RNAN by mixing a local masked branch and a non-local attention mechanism, which are, respectively, in charge of concentrating on extracting more local structures and considering more long-range dependencies in the extracted features. Mei \textit{et al.}~\cite{mei2020image} propose the CSNLN by integrating a Cross-Scale Non-Local prior with local and in-scale non-local priors using a recurrent neural network, which can efficiently explore the existing cross-scale feature similarities in images. Xia \textit{et al.}~\cite{xia2022efficient} propose an efficient non-local attention module by using the kernel function of approximation and the associative law of matrix multiplication, which successfully achieves comparable performance compared to that of the previous non-local attention module while requires only linear computation and space complexity with respect to the LR size. However, these models are still incapable of adequately and comprehensively compensate for the shortcomings of CNNs in establishing long-range dependencies.   

\noindent \textbf{Transformer-based Models:} 
Inspired by the significant performance of the vision Transformer~\cite{vaswani2017attention,devlin2018bert,wang2019glue}, it has also been applied to the SISR field~\cite{yang2020learning,chen2021pre,cao2021video,wang2022uformer,liang2021swinir}. For example, Chen~\textit{et al.} propose the image processing Transformer (IPT)~\cite{chen2021pre} model for various image restoration tasks based on a pre-trained standard Transformer~\cite{vaswani2017attention}. Recently, to capture local relationships, researchers begin to introduce convolutions to Transformers by integrating the vision Transformer module with convolution~\cite{liang2021swinir,zhang2022efficient,huang2023deep,lu2022transformer,fang2022hybrid}. For example, Liang~\textit{et al.}~\cite{liang2021swinir} propose the Swin Transformer-based image resolution model (SwinIR) by combining CNN and Transformer and achieves superior performance while maintaining computational efficiency. Huang~\textit{et al.} propose DGSM-Swin~\cite{huang2023deep} by introducing a learned Gaussian Scale Mixture (GSM) prior into the Swin Transformer. In addition to classic performance-oriented SISRs, hybrid architectures have also emerged in the field of light-weigh SISR~\cite{lu2022transformer,fang2022hybrid,conde2023swin2sr}. For example, Lu~\textit{et al.} propose a novel Efficient Super-Resolution Transformer (ESRT) ~\cite{lu2022transformer}, which integrates a lightweight CNN backbone and a lightweight Transformer backbone to achieve a small GPU memory footprint using an efficient multi-head attention. Fang~\textit{et al.} proposed a Hybrid Network of CNN and Transformer (HNCT)~\cite{fang2022hybrid} for lightweight image SISR, which can exploit both local and non-local priors by integrating CNN and Transformer. Although Transformer-based SISRs have achieved impressive results, most existing Transformers divide images into the same number of fixed-size patches, which may not be ideal for restoring patches with different levels of texture richness. Besides, the position encodings used in most existing Transformer are predefined and treat the positional information of different tokens equally.

\section{Methodology}
\label{sec_HIPA}
\subsection{Issues and Motivations}
\label{motivation}
\begin{figure}[!htb]
	\centering
	\includegraphics[width=0.42\textwidth]{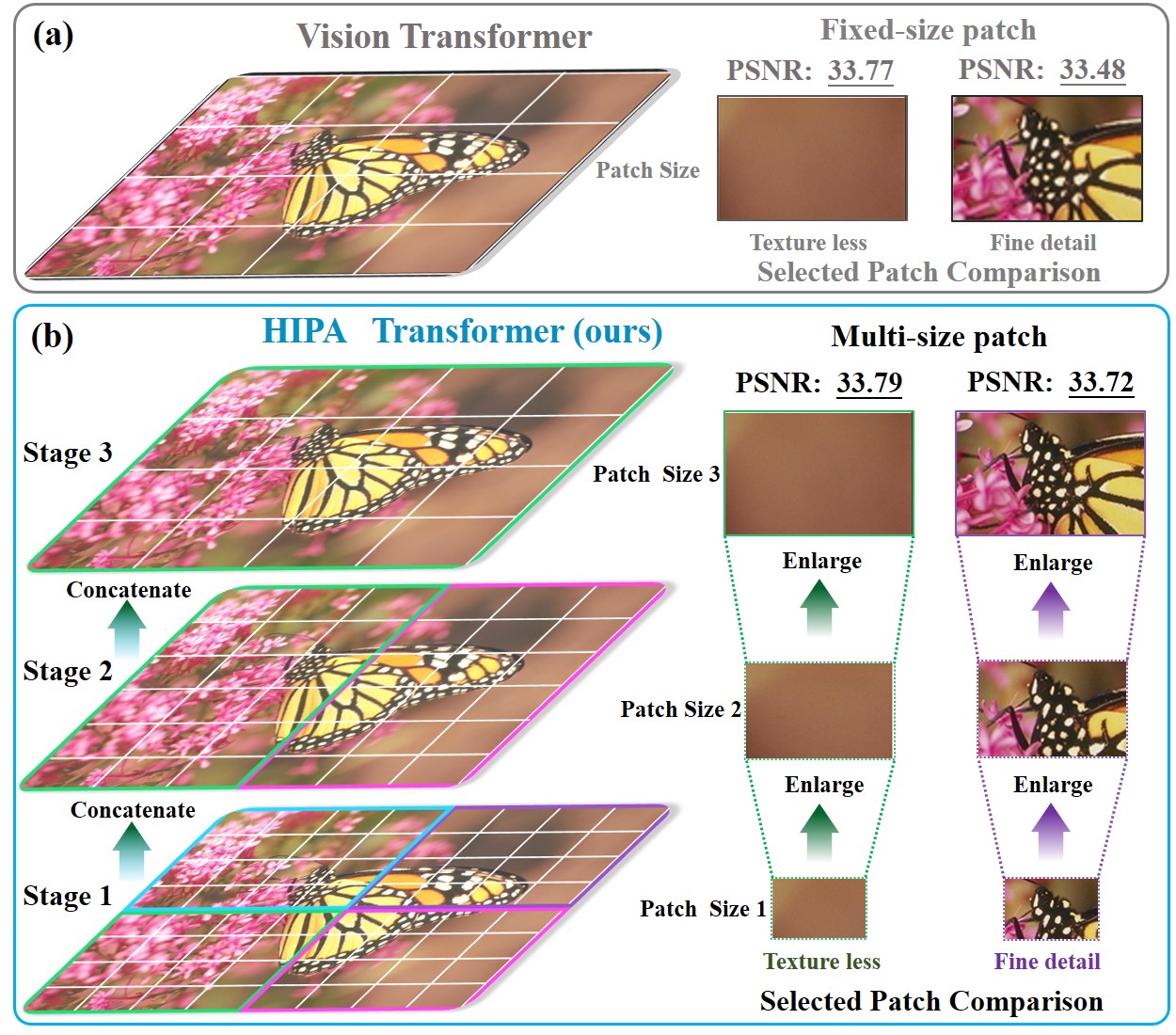}
	\caption{Illustration of comparison between (a) the existing vision Transformer with fixed-size patch and (b) our HIPA Transformer with multi-size patch on regions with different texture richness.}
	\vspace{-4 mm}
	\label{fig_HIPAvsVIT}
\end{figure}
\begin{figure*}[!htb]
	\centering
	\includegraphics[width=0.83\textwidth]{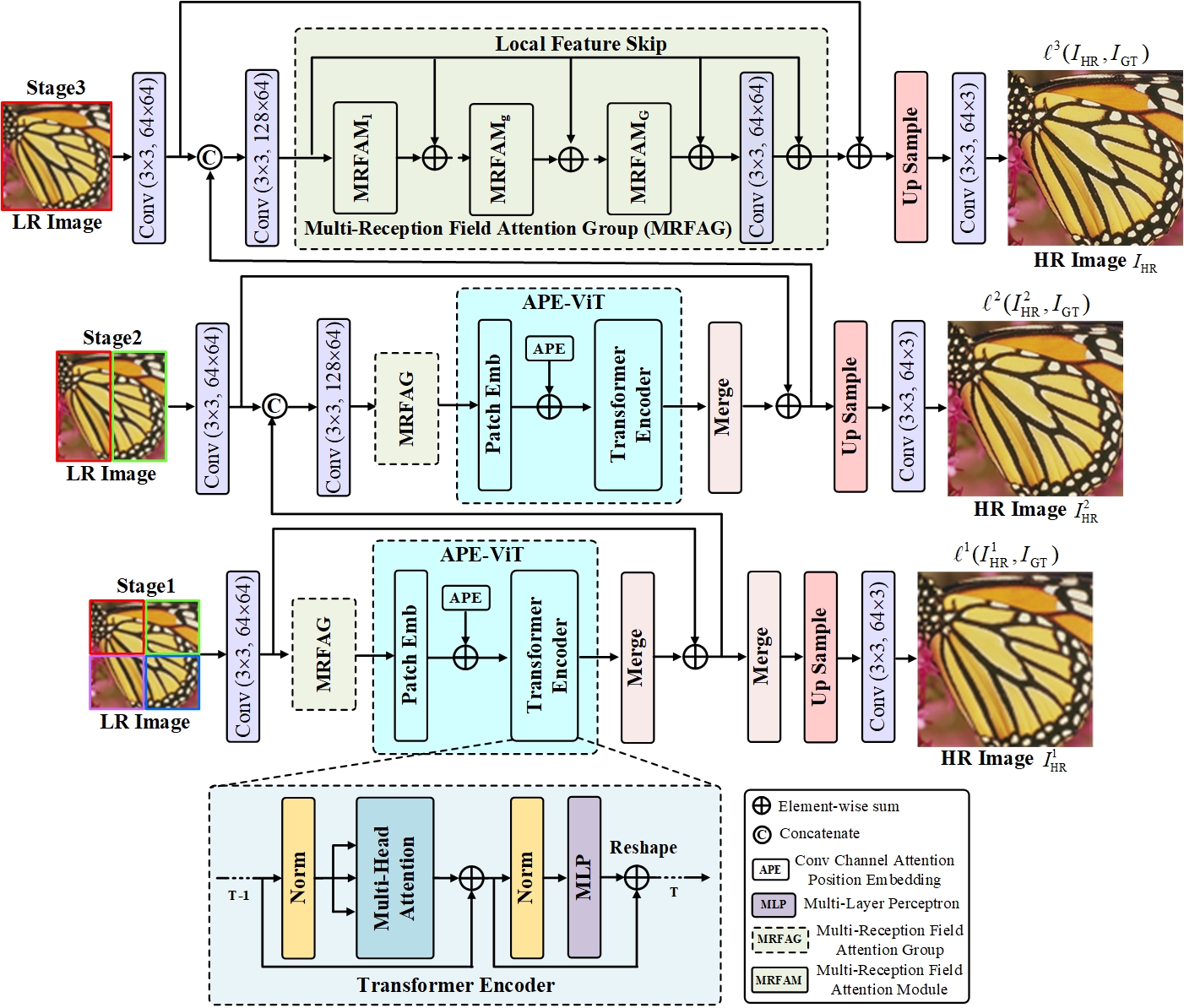}
	\caption{Overall framework of our HIPA method for progressive SISR. Each stage in the proposed model is constructed based on the proposed HIPA block, which consists of two main modules. The first module is a Transformer designed to learn global dependencies between contexts (See Section~\ref{Sec_HIPA} for details). The second module is a multi-receptive field attention group used to exhaustively mine local features contained in the original LR image (See Section~\ref{sec_MRFAG} for details). The Transformers of the first two stages mainly learn the broad contextual information, while the last stage focuses more on learning the desired details. Finally, the training loss is defined based on the summation over all outputs  $\ell^{1}(\cdot)$, $\ell^{2}(\cdot)$, $\ell^{3}(\cdot)$ of different stages to optimize our HIPA.}
	\vspace{-4 mm}
	\label{fig_architecture}
\end{figure*}
As discussed in the contribution part of Section~\ref{sec_Int} and the Transformer-based models part of Section \ref{sec_RW}, using fixed-size patches with varying level of texture richness is suboptimal and can limit the recovery performance of many existing Transformer-based SISR models. To further explain this, we provide an example shown in Fig.~\ref{fig_HIPAvsVIT} that demonstrates the impact of patch size on the ``monarch" image from the Set14 dataset. Fig.~\ref{fig_HIPAvsVIT}(a) shows the fixed-size split of vision Transformer-based methods (left column), in which the input image is split into the same number of fixed-size tokens, a selected textureless background region (middle column) and a butterfly head region with fine detail (right column). Fig.~\ref{fig_HIPAvsVIT}(b) shows the multi-size split of our HIPA Transformer (left column), in which the input image is partitioned into multiple stages, where tokens with small patch sizes are used first and gradually merged with larger patches to form the full resolution, two selected regions (last two columns) the same as that in Fig.~\ref{fig_HIPAvsVIT}(a). From the visual and quantitative comparison of the selected background region (middle column) between using the existing fixed-size patch and our multi-size patch, it can be observed that their visual quality and PSNR values are very similar, which suggests that using a large patch size for textureless region is enough for the network to finish the final restoration. However, from the recovery performance comparison of the selected butterfly head region (right column) between using the existing fixed-size patch and using our multi-size patch, it can be found that their visual quality and PSNR values have a certain gap, which suggests that using a large patch size for region with fine detail is not optimal. In contrast, in this case, using a smaller patch size is more helpful to recover fine details, which is demonstrated by the improved PSNR value using our multi-size patch. From the above discussion, we can summarize that: (1) using fixed-size patches with different texture richness in the whole restoration process is inappropriate, which is the reason that many existing Transformer-based models still exhibit blurry boundaries; (2) using multi-size patches with different richness is helpful to improve the restoration results, which motivates the proposed HIPA Transformer. 

\subsection{Hierarchical Patch Transformer}
\label{Sec_HIPA}
As shown in Fig.~\ref{fig_architecture}, the proposed HIPA consists of three stages to progressively recover the high-resolution (HR) image from its low-resolution (LR) input. The Transformer of the first two stages mainly learn broad contextual information, while the last stage focuses on learning the desired details. Each stage is constructed based on the proposed HIPA block, which mainly consists of two modules: a multi-receptive field attention group and a designed Transformer. To achieve multi-size patch input for the HIPA, we adopt the hierarchical patch partition on the input LR image. Specifically, we first split the LR image into different non-overlapping patches for different stages: four for the first stage, two for the second stage, and the entire LR image for the last stage, and then, gradually integrate intermediate results in the next stage.    

For simplicity, in the notation that follows, $I_{LR}$ and $I_{HR}$ denote the original LR input and the final HR output of the HIPA, respectively. $I_{LR}^{i,j}$ denotes the $j$-th patch at Stage $i$. For example, $I_{LR}^{1,1}$ denotes the $1$-st patch at Stage 1, i.e., the upper left corner patch of Stage 1 input shown in Fig.~\ref{fig_architecture}. 

Following~\cite{lim2017enhanced,zhang2018image}, we also use one convolution layer to extract the shallow feature (SF) $F_{0}^{1,j}$ from the original LR image. For Stage 1:
\begin{equation}
F_{0}^{1,j}=H_{SF}(I_{LR}^{1,j})~~(j=1,2,3,4),
\end{equation}
where $H_{SF}$ denotes the convolution operation. Then, the extracted shallow feature is input to the proposed HIPA block to further extract deep features:
\begin{equation}
F_{HIPA}^{1,j}=H_{HIPA}(F_{0}^{1,j})~~(j=1,2,3,4),
\end{equation}
where $H_{HIPA}$ denotes the proposed HIPA block. After stitching $F_{HIPA}^{1,1}$ with $F_{HIPA}^{1,3}$ and stitching $F_{HIPA}^{1,2}$ with $F_{HIPA}^{1,4}$ using concatenate operation, dubbed vertical stitching, we obtain the output features of Stage 1, which are then concatenated with the shallow features of Stage 2 as shown in Fig.~\ref{fig_architecture}:
\begin{equation}
\begin{aligned}
F_{Sti}^{1,1}=&H_{Sti}(F_{HIPA}^{1,1},F_{HIPA}^{1,3})+H_{Sti}(F_{0}^{1,1},F_{0}^{1,3}),\\
F_{Sti}^{1,2}=&H_{Sti}(F_{HIPA}^{1,2},F_{HIPA}^{1,4})+H_{Sti}(F_{0}^{1,2},F_{0}^{1,4}),
\end{aligned}
\end{equation}
where $H_{Sti}$ denotes the stitch using the concatenate operation. We utilize vertical stitching for sub-patches rather than horizontal stitching, i.e., stitching $F_{HIPA}^{1,1}$ with $F_{HIPA}^{1,2}$ and stitching $F_{HIPA}^{1,3}$ with $F_{HIPA}^{1,4}$ using the concatenate operation. Although we also investigated horizontal stitching, it did not yield significant differences. Finally, the recovered HR image of Stage 1: $I_{HR}^1$, is obtained by further stitching $F_{Sti}^{1,1}$ and $F_{Sti}^{1,2}$ using concatenate operation, and then successively input the stitched result into an upscale module and a reconstruction module (i.e., one convolution layer) as follows:
\begin{equation}
\begin{split}
I_{HR}^1=H_{Rec}(H_{UP}(H_{Sti}(F_{Sti}^{1,1},F_{Sti}^{1,2}))),
\end{split}
\end{equation}
where $H_{UP}$ and $H_{Rec}$ denote the upscale and reconstruction module, respectively. 

For Stage 2 and Stage 3, the extracted shallow features $F_0^{2,j}~~(j=1,2)$ and $F_0^{3,1}$ need to be first concatenated with the output features of the upper stage, which is then input into the next operation similar to Stage 1. Finally, the recovered HR images of Stage 2: $I_{HR}^2$ and Stage 3: $I_{HR}$ can be obtained.  As shown in Fig.~\ref{fig_architecture}, the predictions of the three stages are gradually improved. For example, the prediction of Stage 2 is the refinement of Stage 1. With the multi-stage refinement, image regions with high spatial frequency are gradually recovered.

Finally, the proposed HIPA is trained using a training loss, which is the sum over all the outputs of $I_{HR}^1$ (Stage 1), $I_{HR}^2$ (Stage 2) and $I_{HR}$ (Stage 3):
\begin{equation}
\label{eq_loss}
\begin{aligned}
L(\varTheta)=&\ell^{1}(I_{HR}^1, I_{GT})+\ell^{2}(I_{HR}^2, I_{GT\_T})+\ell^{3}(I_{HR}, I_{GT}),
\end{aligned}
\end{equation}
where $\varTheta$ denotes the parameter set of the proposed network. $\ell^{1}(\cdot)$, $\ell^{2}(\cdot)$ and $\ell^{3}(\cdot)$, respectively, stand for the loss of Stage 1, Stage 2 and Stage3. This work also uses the $L_1$ loss following previous work for the sake of fairness. $I_{GT}$ denotes the ground-truth HR image. 

\subsection{Attention-based Position Encoding} 
\label{sec_Vit}
\begin{figure}[!htb]
	\centering
	\includegraphics[width=0.48\textwidth]{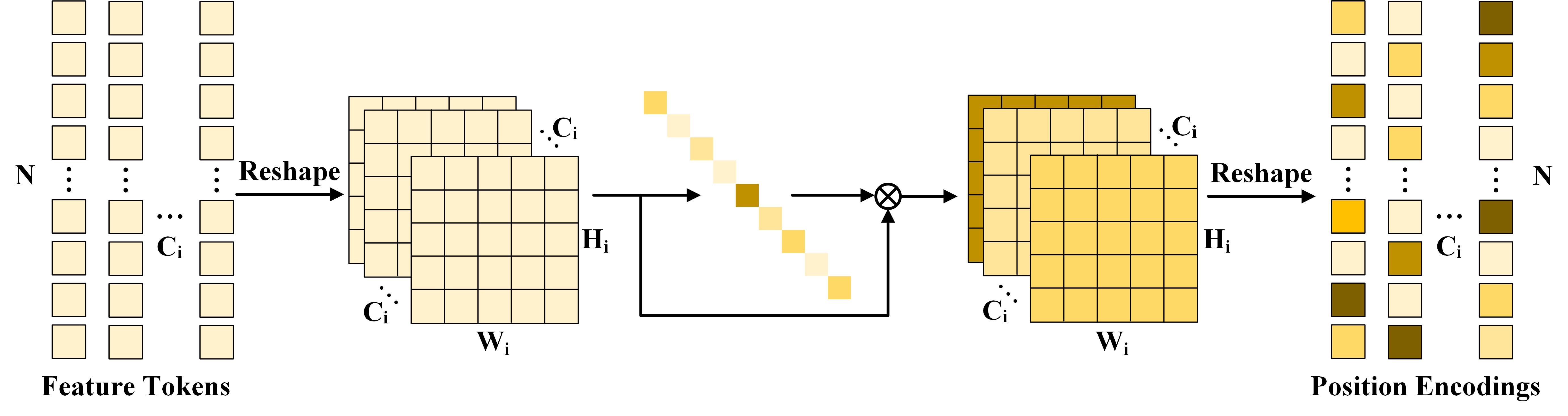}
	\caption{Illustration of attention-based position encoding (APE). $N=\frac{H_i\times W_i}{P_i^2}$ denotes the number of tokens. $W$, $H$ and $C$, respectively, denote the width, height and channel number of the feature map.}
	\vspace{-2 mm}
	\label{fig_ape}
\end{figure}
To incorporate the order of the token sequence, position encodings are usually adopted in Transformers~\cite{chu2021conditional}. However, the original position embedding of ViT is pre-defined and independent of input tokens. When an input LR image with a new size is input, the number of patches will be different from that before and the learned position embedding will be mismatched with the new size. So, the input image with a new size has to be first interpolated to the desired size, which not only reduces the overall performance of the ViT but also seriously limits its application. To address the above issue, Chu \textit{et al.}~\cite{chu2021conditional} propose a conditional position encoding (CPE) by introducing a 2-D convolution to embed position encoding, which can easily generalize to an input LR image with a new input size. However, the CPE treats all input tokens equally and may neglect the dependencies among them. To address this, we propose a new attention-based position encoding (APE) method by introducing attention into position embedding to let the Transformer focus on important tokens. Specifically, the patch embedding module first reshapes the extracted feature $F_{MRFAG}^{i,j}\in R^{H_i\times W_i\times C_i}$ into a number of flattened 2D patches $x_p \in R^{\frac{H_i\times W_i}{P^2}\times P^2\times C_i}$ by partitioning the input into non-overlapping $P_i\times P_i$ patches where $(H_i, W_i)$, $C_i$, $\frac{H_i\times W_i}{P_i^2}$ and $P_i$, respectively, denote the resolution of Stage $i$ input, the number of channel, the number of patches and the patch size. Then, as shown in Fig.~\ref{fig_ape}, the flattened feature tokens are reshaped to the 2D image space. In the 2D image space, a convolution and a channel attention are applied to produce the final position encoding. With the help of attention, the final position encoding can let the network focus on the important tokens.

\subsection{Multi-Receptive Field Attention Group}
\label{sec_MRFAG}
\begin{figure}[!htb]
	\centering
	\includegraphics[width=0.44\textwidth]{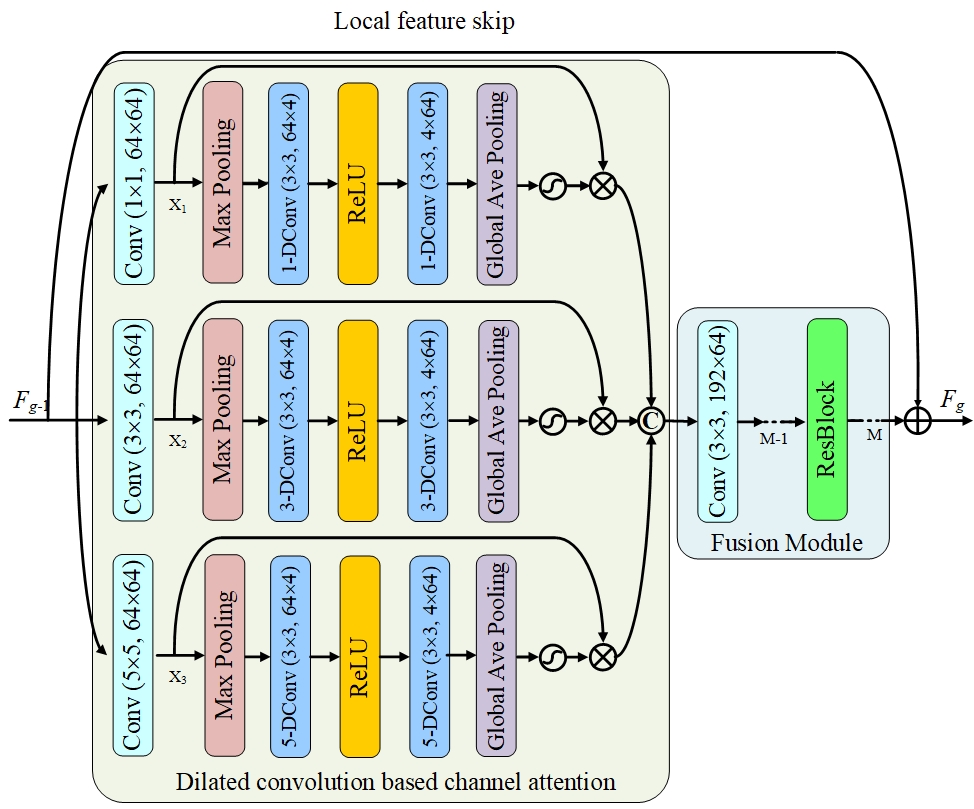}
	\caption{The architecture of the multi-receptive field attention module (MRFAM). Note the numbers $(kernel\times kernel, input \times output)$ in Conv and Dconv denote kernel size and input and output feature map number. $i$-DConv denotes dilated convolution with dilation factor $i$.}
	\vspace{-4 mm}
	\label{fig_mrfag}
\end{figure}
We now show our MRFAG, which mainly consists of $G$ multi-receptive field attention modules (MRFAMs) as shown in Fig.~\ref{fig_mrfag}. Each MRFAM consists of three dilated convolution based channel attention connected in parallel, a fusion module and a local feature skip (LFS). It has been verified that although increasing the depth and the filter size of the CNN can, respectively, enlarge the receptive field and extract more information contained in the low-quality image, it not only introduces more parameters but also increases the computational complexity~\cite{zhang2017learning}. Thus, we propose the dilated convolution based channel attention to enlarge the receptive field of the networks, which is the most significant difference between ours and the Squeeze-and-Excitation network (SE)~\cite{hu2018squeeze}. 

Specifically, for each dilated convolution based channel attention shown in Fig.~\ref{fig_mrfag}, denote $X_i=[x_{i,1},\cdots, x_{i,c}, \cdots, x_{i,C}]$ to be the input, which contains $C$ 2D feature map $x_{i,c} \in R^{H \times W}$, where $H$ and $W$, respectively, are the height and width of the feature map. Firstly, by shrinking the extracted features using max pooling, the output feature  $Z_i=[z_{i,1},\cdots, z_{i,c}, \cdots, z_{i,C}]$ of each branch can be obtained, where $z_{i,c}\in R^{\frac{H}{Stride} \times \frac{W}{Stride}}$ denotes the output feature. Then, two dilated convolution layers and an activation function are applied to fully exploit feature dependencies from the aggregated information. Finally, the sigmoid function is adopted as the activation function:
\begin{equation}
s_{i,c}=f(H_{GPL}(W_{U}\delta(W_{D}z_{i,c}))),
\end{equation}
where $f(\cdot)$, $H_{GPL}(\cdot)$ and $\delta(\cdot)$, respectively, stand for the sigmoid function, the global average pooling and the ReLU function. $W_{D}$ is the weight set of the first dilated convolution layer in the channel attention shown in Fig.~\ref{fig_mrfag}, which plays the role of downscaling with a reduction ratio $\gamma$ (we set $\gamma=16$). After the ReLU function, the low-dimension feature is then upsampled with ratio $\gamma$ by the second dilated convolution layer. $W_U$ denotes its weight set. The channel statistics $s$ can be obtained to rescale the input $x_{i,c}$:
\begin{equation}
\hat x_{i,c}=s_{i,c}\cdot x_{i,c},
\end{equation}
where $s_{i,c}$ and $x_{i,c}$ denote, respectively, the scaling factor and feature maps of the $c$-th channel.

Besides, we introduce the LFS connection to ensure stability in training the network and to bypass redundant features in the low-quality image. The final output of MRFAG is obtained as
\begin{equation}
F_{MRFAG}=F_0^{i,j}+\omega_{LFS}(F_{MRFAM_{G}}),
\end{equation}
where $\omega_{LFS}$ denotes the weight of the convolution layer at the tail of MRFAG. $F_{MRFAG}$ and $F_{MRFAM_{G}}$, respectively, denote the output of MRFAG and the $G$-th output of MRFAM.
\begin{figure}[!htb]
	\centering
	\includegraphics[width=0.44\textwidth]{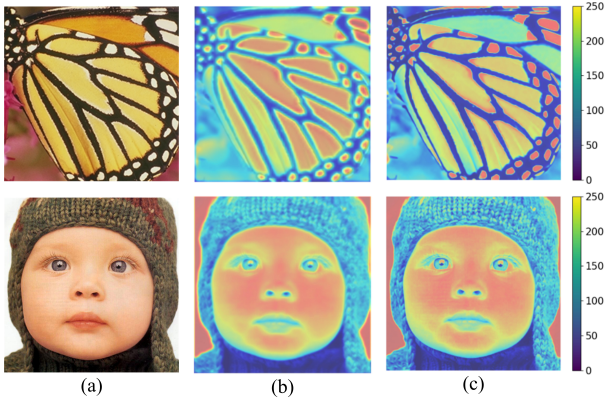}
	\caption{Class activate map (CAM) comparison. (a) Input images. (b) CAM before using MRFAG. (c) CAM after using MRFAG.}
	\vspace{-4 mm}
	\label{fig_mrfag_CAM}
\end{figure}

Fig.~\ref{fig_mrfag_CAM} shows a comparison of the class active map (CAM) before and after using the MRFAG. It can be found that the CAM after using MRFAG becomes sharper than before, validating that the MRFAG has a better capability of recovering high-frequency signals. This also enables the network to focus more on recovering textures and details. 

\subsection{Discussions}
Below, we discuss the significant differences between our HIPA Transformer and two most closely related Transformers: IPT~\cite{chen2021pre} and SwinIR~\cite{liang2021swinir}.

\textbf{Differences to the IPT model:}
Based on a pre-trained standard Transformer~\cite{vaswani2017attention}, Chen~\textit{et al.} propose the image processing Transformer (IPT)~\cite{chen2021pre} model for image restoration tasks, which achieves superior performance than most CNN-based SISR methods. Although it is a Transformer-based SISR method similar to our method, there are three significant differences between the IPT and our HIPA Transformer: (i) the IPT model uses a pre-trained Transformer, which means that when we apply it for image super-resolution, we need to first pre-train the Transformer using large labeled datasets and then fine-tune the whole network. As a result, its performance is limited by the lack of sufficient labeled samples for training. In contrast, our HIPA Transformer is an end-to-end network, which successfully avoids the tedious pre-training and fine-tuning process; (ii) IPT uses fixed-size patches for all input tokens with different richness, which is not optimal and limits its performance as discussed in Section~\ref*{motivation}, while our HIPA Transformer uses multi-size patches for tokens with different richness, e.g., using a smaller patch for areas with fine details and a large patch for textureless regions; (iii) The IPT uses Transformer to extract features and to construct long-range dependencies, while our HIPA Transformer is a hybrid architecture combining CNN and Transformer, which can fully utilize the advantage of CNN in local feature extraction and the advantage of Transformer in establishing long-range dependencies. More comparisons of experimental results are shown in Section~\ref{sec_compar}. 
\begin{table*}[ht]
	\centering
	\scriptsize
	\renewcommand{\arraystretch}{0.9} 
	\caption{Quantitative comparisons with state-of-the-art performance-oriented SISR methods on five benchmark datasets for scale factor $\times2$, $\times3$ and $\times4$. The best results are \textbf{highlighted} in \textcolor{red}{red} and the second best in \textcolor{blue}{blue}.}
	\label{tab_ComSR}
	\begin{tabular}{ p{1.8cm} | p{0.71cm}<{\centering} | p{0.71cm}<{\centering} | p{0.71cm}<{\centering} | p{0.71cm}<{\centering} | p{0.71cm}<{\centering} | p{0.71cm}<{\centering} | p{0.71cm}<{\centering} | p{0.71cm}<{\centering} | p{0.71cm}<{\centering} | p{0.71cm}<{\centering} | p{0.71cm}<{\centering} | p{0.71cm}<{\centering}  }
		\hline
		\multicolumn{1}{c|}{\multirow{2}{*}{Methods}} & \multicolumn{1}{c|}{\multirow{2}{*}{Scale}} & \multicolumn{1}{c|}{\multirow{2}{*}{Year}} &
		\multicolumn{2}{c|}{Set5} & \multicolumn{2}{c|}{Set14} & \multicolumn{2}{c|}{B100}  & \multicolumn{2}{c|}{Urban100} &
		\multicolumn{2}{c}{Manga109} \\ \cline{4-13}
		&  & & PSNR      & SSIM  & PSNR   & SSIM  & PSNR   & SSIM  & PSNR   & SSIM  & PSNR   & SSIM \\\hline\hline                              
		SRMD~\cite{zhang2018learning}  & $\times2$ & 2018 & 37.79 & 0.9601 & 33.32 & 0.9159 & 32.05 & 0.8985 & 31.33 & 0.9204 & 38.07 & 0.9761 \\
		DBPN~\cite{haris2018deep}      & $\times2$ & 2018 & 38.09 & 0.9600 & 33.85 & 0.9190 & 32.27 & 0.9000 & 32.55 & 0.9324 & 38.89 & 0.9775 \\
		RDN~\cite{zhang2018residual}   & $\times2$ & 2018 & 38.24 & 0.9614 & 34.01 & 0.9212 & 32.34 & 0.9017 & 32.89 & 0.9353 & 39.18 & 0.9780 \\
		MSRN~\cite{li2018multi}        & $\times2$ & 2018 & 38.08 & 0.9605 & 33.74 & 0.9170 & 32.23 & 0.9013 & 32.22 & 0.9326 & 38.82 & 0.9768 \\
		RCAN~\cite{zhang2018image}     & $\times2$ & 2018 & 38.27 & 0.9614 & 34.12 & 0.9216 & 32.41 & 0.9027 & 33.34 & 0.9384 & 39.44 & 0.9786 \\
		SRFBN~\cite{li2019feedback}    & $\times2$ & 2019 & 38.11 & 0.9609 & 33.82 & 0.9196 & 32.29 & 0.9010 & 32.62 & 0.9328 & 39.08 & 0.9779 \\
		SAN~\cite{dai2019second}       & $\times2$ & 2019 & 38.31 & 0.9620 & 34.07 & 0.9213 & 32.42 & 0.9028 & 33.10 & 0.9370 & 39.32 & 0.9792 \\
		CSNLN~\cite{mei2020image}      & $\times2$ & 2020 & 38.28 & 0.9616 & 34.12 & 0.9223 & 32.40 & 0.9024 & 33.25 & 0.9386 & 39.37 & 0.9785 \\
		HAN~\cite{niu2020single}       & $\times2$ & 2020 & 38.27 & 0.9614 & 34.16 & 0.9217 & 32.41 & 0.9027 & 33.35 & 0.9385 & 39.46 & 0.9785 \\
		NSR~\cite{fan2020neural}       & $\times2$ & 2020 & 38.23 & 0.9614 & 33.94 & 0.9203 & 32.34 & 0.9020 & 33.02 & 0.9367 & 39.31 & 0.9782 \\
		IGNN~\cite{zhou2020cross}      & $\times2$ & 2020 & 38.24 & 0.9613 & 34.07 & 0.9217 & 32.41 & 0.9025 & 33.23 & 0.9386 & 39.35 & 0.9786 \\
		RFANet~\cite{liu2020residual}  & $\times2$ & 2020 & 38.26 & 0.9615 & 34.16 & 0.9220 & 32.41 & 0.9026 & 33.33 & 0.9389 & 39.44 & 0.9783 \\
		NLSN~\cite{mei2021image}       & $\times2$ & 2021 & 38.34 & 0.9618 & 34.08 & 0.9231 & 32.43 & 0.9027 & 33.42 & 0.9394 & 39.59 & 0.9789 \\
		SwinIR~\cite{liang2021swinir}  & $\times2$ & 2021 & 38.35 & 0.9620 & 34.14 & 0.9227 & 32.44 & 0.9030 & 33.40 & 0.9393 & 39.60 & 0.9792 \\
		TDPN~\cite{cai2022tdpn}        & $\times2$ & 2022 & 38.31 & 0.9621 & 34.16 & 0.9225 & 32.52 & 0.9045 & 33.36 & 0.9386 & 39.57 & 0.9795 \\
		ELAN~\cite{zhang2022efficient} & $\times2$ & 2022 & 38.36 & 0.9620 & 34.20 & 0.9228 & 32.45 & 0.9030 & 33.44 & 0.9391 & 39.62 & 0.9793 \\
		DGSM-Swin~\cite{huang2023deep} & $\times2$ & 2023 & 38.24 & 0.9615 & 33.93 & 0.9217 & 32.36 & 0.9019 & 32.95 & 0.9442 & 39.31 & 0.9783 \\ \hline
		
		HIPA(ours)                     & $\times2$ & 2023 & \textcolor{blue}{38.38} & \textcolor{blue}{0.9621} & \textcolor{blue}{34.25} & \textcolor{blue}{0.9235} & \textcolor{blue}{32.48} & \textcolor{blue}{0.9033} & \textcolor{blue}{33.50} & \textcolor{blue}{0.9400} & \textcolor{blue}{39.75} & \textcolor{blue}{0.9794} \\
		HIPA+(ours)                    & $\times2$ & 2023 & \textcolor{red}{38.41} & \textcolor{red}{0.9623} & \textcolor{red}{34.30} & \textcolor{red}{0.9238} & \textcolor{red}{32.51} & \textcolor{red}{0.9036} & \textcolor{red}{33.57} & \textcolor{red}{0.9409} & \textcolor{red}{39.81} & \textcolor{red}{0.9795} \\
		\hline\hline
		SRMD~\cite{zhang2018learning}  & $\times3$ & 2018 & 34.12 & 0.9254 & 30.04 & 0.8382 & 28.97 & 0.8025 & 27.57 & 0.8398 & 33.00 & 0.9403 \\
		RDN~\cite{zhang2018residual}   & $\times3$ & 2018 & 34.71 & 0.9296 & 30.57 & 0.8468 & 29.26 & 0.8093 & 28.80 & 0.8653 & 34.13 & 0.9484 \\
		MSRN~\cite{li2018multi}        & $\times3$ & 2018 & 34.38 & 0.9262 & 30.34 & 0.8395 & 29.08 & 0.8041 & 28.08 & 0.8554 & 33.44 & 0.9427\\
		RCAN~\cite{zhang2018image}     & $\times3$ & 2018 & 34.74 & 0.9299 & 30.65 & 0.8482 & 29.32 & 0.8111 & 29.09 & 0.8702 & 34.44 & 0.9499 \\
		SRFBN~\cite{li2019feedback}    & $\times3$ & 2019 & 34.70 & 0.9292 & 30.51 & 0.8461 & 29.24 & 0.8084 & 28.73 & 0.8641 & 34.18 & 0.9481 \\
		SAN~\cite{dai2019second}       & $\times3$ & 2019 & 34.75 & 0.9300 & 30.59 & 0.8476 & 29.33 & 0.8112 & 28.93 & 0.8671 & 34.30 & 0.9494 \\
		CSNLN~\cite{mei2020image}      & $\times3$ & 2020 & 34.74 & 0.9300 & 30.66 & 0.8482 & 29.33 & 0.8105 & 29.13 & 0.8712 & 34.45 & 0.9502 \\
		HAN~\cite{niu2020single}       & $\times3$ & 2020 & 34.75 & 0.9299 & 30.67 & 0.8483 & 29.32 & 0.8110 & 29.10 & 0.8705 & 34.48 & 0.9500 \\
		NSR~\cite{fan2020neural}       & $\times3$ & 2020 & 34.62 & 0.9289 & 30.57 & 0.8475 & 29.26 & 0.8100 & 28.83 & 0.8663 & 34.27 & 0.9484 \\
		IGNN~\cite{zhou2020cross}      & $\times3$ & 2020 & 34.72 & 0.9298 & 30.66 & 0.8484 & 29.31 & 0.8105 & 29.03 & 0.8696 & 34.39 & 0.9496 \\
		RFANet~\cite{liu2020residual}  & $\times3$ & 2020 & 34.79 & 0.9300 & 30.67 & 0.8487 & 29.34 & 0.8115 & 29.15 & 0.8720 & 34.59 & 0.9506 \\
		NLSN~\cite{mei2021image}       & $\times3$ & 2021 & 34.85 & 0.9306 & 30.70 & 0.8485 & 29.34 & 0.8117 & 29.25 & 0.8726 & 34.57 & 0.9508 \\
		SwinIR~\cite{liang2021swinir}  & $\times3$ & 2021 & 34.89 & 0.9312 & 30.77 & 0.8503 & 29.37 & 0.8124 & 29.29 & 0.8744 & 34.74 & 0.9518 \\
		TDPN~\cite{cai2022tdpn}        & $\times3$ & 2022 & 34.86 & 0.9312 & 30.79 & 0.8501 & 29.45 & 0.8126 & 29.26 & 0.8724 & 34.48 & 0.9508 \\
		ELAN~\cite{zhang2022efficient} & $\times3$ & 2022 & 34.90 & 0.9313 & 30.80 & 0.8504 & 29.38 & 0.8124 & 29.32 & 0.8745 & 34.73 & 0.9517 \\
		DGSM-Swin~\cite{huang2023deep} & $\times3$ & 2023 & 34.77 & 0.9300 & 30.65 & 0.8490 & 29.29 & 0.8109 & 28.93 & 0.8684 & 34.30 & 0.9498 \\\hline
		HIPA(ours)                     & $\times3$ & 2023 & \textcolor{blue}{34.95} & \textcolor{blue}{0.9318} & \textcolor{blue}{30.84} & \textcolor{blue}{0.8515} & \textcolor{blue}{29.45} & \textcolor{blue}{0.8140} & \textcolor{blue}{29.41} & \textcolor{blue}{0.8760} & \textcolor{blue}{34.88} & \textcolor{blue}{0.9521} \\
		HIPA+(ours)                    & $\times3$ & 2023 & \textcolor{red}{35.01} & \textcolor{red}{0.9320} & \textcolor{red}{30.90} & \textcolor{red}{0.8530} & \textcolor{red}{29.49} & \textcolor{red}{0.8151} & \textcolor{red}{29.50} & \textcolor{red}{0.8784} & \textcolor{red}{34.96} & \textcolor{red}{0.9528} \\
		\hline \hline
		SRMD~\cite{zhang2018learning}  & $\times4$ & 2018 & 31.96 & 0.8925 & 28.35 & 0.7787 & 27.49 & 0.7337 & 25.68 & 0.7731 & 30.09 & 0.9024 \\
		DBPN~\cite{haris2018deep}      & $\times4$ & 2018 & 32.47 & 0.8980 & 28.82 & 0.7860 & 27.72 & 0.7400 & 26.38 & 0.7946 & 30.91 & 0.9137 \\
		RDN~\cite{zhang2018residual}   & $\times4$ & 2018 & 32.47 & 0.8990 & 28.81 & 0.7871 & 27.72 & 0.7419 & 26.61 & 0.8028 & 31.00 & 0.9151 \\
		MSRN~\cite{li2018multi}        & $\times4$ & 2018 & 32.07 & 0.8903 & 28.60 & 0.7751 & 27.52 & 0.7273 & 26.04 & 0.7896 & 30.17 & 0.9034 \\
		RCAN~\cite{zhang2018image}     & $\times4$ & 2018 & 32.63 & 0.9002 & 28.87 & 0.7889 & 27.77 & 0.7436 & 26.83 & 0.8087 & 31.22 & 0.9173 \\
		SRFBN~\cite{li2019feedback}    & $\times4$ & 2019 & 32.47 & 0.8983 & 28.81 & 0.7868 & 27.72 & 0.7409 & 26.60 & 0.8015 & 31.15 & 0.9160 \\
		SAN~\cite{dai2019second}       & $\times4$ & 2019 & 32.64 & 0.9003 & 28.92 & 0.7888 & 27.78 & 0.7436 & 26.79 & 0.8068 & 31.18 & 0.9169 \\
		CSNLN~\cite{mei2020image}      & $\times4$ & 2020 & 32.68 & 0.9004 & 28.95 & 0.7888 & 27.80 & 0.7439 & 27.22 & 0.8168 & 31.43 & 0.9201 \\
		HAN~\cite{niu2020single}       & $\times4$ & 2020 & 32.64 & 0.9002 & 28.90 & 0.7890 & 27.80 & 0.7442 & 26.85 & 0.8094 & 31.42 & 0.9177 \\
		NSR~\cite{fan2020neural}       & $\times4$ & 2020 & 32.55 & 0.8987 & 28.79 & 0.7876 & 27.72 & 0.7414 & 26.61 & 0.8025 & 31.10 & 0.9145 \\
		IGNN~\cite{zhou2020cross}      & $\times4$ & 2020 & 32.57 & 0.8998 & 28.85 & 0.7891 & 27.77 & 0.7434 & 26.84 & 0.8090 & 31.28 & 0.9182 \\
		RFANet~\cite{liu2020residual}  & $\times4$ & 2020 & 32.66 & 0.9004 & 28.88 & 0.7894 & 27.79 & 0.7442 & 26.92 & 0.8112 & 31.41 & 0.9187 \\
		NLSN~\cite{mei2021image}       & $\times4$ & 2021 & 32.59 & 0.9000 & 28.87 & 0.7891 & 27.78 & 0.7444 & 26.96 & 0.8109 & 31.27 & 0.9184 \\
		SwinIR~\cite{liang2021swinir}  & $\times4$ & 2021 & 32.72 & 0.9021 & 28.94 & 0.7914 & 27.83 & 0.7459 & 27.07 & 0.8164 & 31.67 & 0.9226\\
		TDPN~\cite{cai2022tdpn}        & $\times4$ & 2022 & 32.69 & 0.9005 & 29.01 & 0.7943 & 27.93 & 0.7460 & 27.24 & 0.8171 & 31.58 & 0.9218 \\
		ELAN~\cite{zhang2022efficient} & $\times4$ & 2022 & 32.75 & 0.9022 & 28.96 & 0.7914 & 27.83 & 0.7459 & 27.13 & 0.8167 & 31.68 & 0.9226\\
		DGSM-Swin~\cite{huang2023deep} & $\times4$ & 2023 & 32.61 & 0.9005 & 28.91 & 0.7903 & 27.78 & 0.7445 & 26.73 & 0.8068 & 31.25 & 0.9193\\	\hline
		HIPA(ours)                     & $\times4$ & 2023 & \textcolor{blue}{32.78} & \textcolor{blue}{0.9025} & \textcolor{blue}{29.07} & \textcolor{blue}{0.7935} & \textcolor{blue}{27.90} & \textcolor{blue}{0.7479} & \textcolor{blue}{27.27} & \textcolor{blue}{0.8191} & \textcolor{blue}{31.83} & \textcolor{blue}{0.9365} \\
		HIPA+(ours)                    & $\times4$ & 2023 & \textcolor{red}{32.84} & \textcolor{red}{0.9034} & \textcolor{red}{29.14} & \textcolor{red}{0.7955} & \textcolor{red}{27.98} & \textcolor{red}{0.7490} & \textcolor{red}{27.31} & \textcolor{red}{0.8214} & \textcolor{red}{31.91} & \textcolor{red}{0.9438} \\
		\hline
	\end{tabular}
\end{table*}

\textbf{ Differences to the SwinIR model:}
Liang~\textit{et al.} propose the SwinIR~\cite{liang2021swinir} by combining CNN with the Swin Transformer~\cite{liu2021swin} into one network, which is also a hybrid architecture similar to our HIPA Transformer. Our key distinctions from it are summarized as follows: (i) The SwinIR uses a plain concatenation of CNN and Transformer with a fixed patch size, while we use a multi-stage model that divides the input into different blocks and aggregates them from small to large by alternating CNN and Transformer, which not only explicitly enables feature aggregation at multiple resolution but also adaptively learns patch-aware features for different image regions; (ii) The local-resolution input tokens contain abundant information for SISR, however, the SwinIR treats all the input tokens equally and hence, limits its representation ability. In contrast, we design a novel attention-based encoding method to focus on the important tokens and to improve its performance for regions with fine details; (iii) The CNN used in the SwinIR model detects local image features using the same scale, treats all LR image features equally, and neglects the dependencies among them. In contrast, our HIPA proposes a multi-receptive field attention module, which lets the proposed network know where to pay more attention and to sufficiently extract local features of LR images. The experimental results, which demonstrate the advantages of our method, are shown in Section~\ref{sec_compar}. 


\section{Experiments}
\label{sec_Exp}
\subsection{Settings}
\label{sec_setting}
\noindent \textbf{Datasets:} Following previous works~\cite{zhang2018image,dai2019second,mei2020image,niu2020single}, we also choose DIV2K~\cite{agustsson2017ntire} as our training dataset, which contains 800 training images and 100 validation images. For testing, we select the standard public datasets: Set5~\cite{bevilacqua2012low}, Set14~\cite{zeyde2010single}, B100~\cite{martin2001database}, Urban100~\cite{huang2015single}, and Manga109~\cite{matsui2017sketch} as our test datasets. All degraded datasets are obtained by the bicubic interpolation model. 
\begin{figure*}[ht]
	\centering
	\scriptsize
	\includegraphics[width=0.88\textwidth]{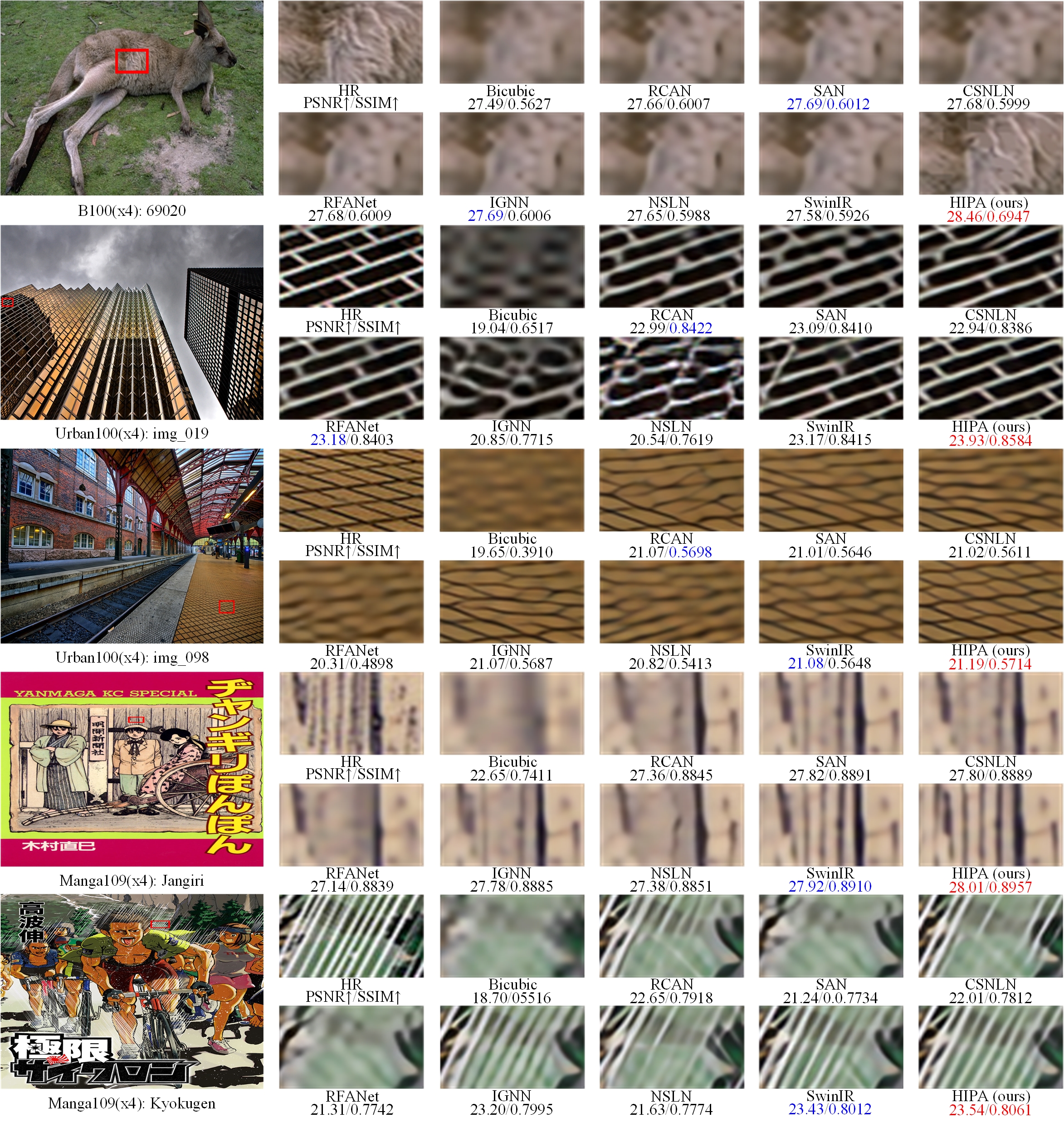}
	\caption{Visual comparisons with state-of-the-art SISR methods for $4\times$ SR on the B100, the Urban100 and the Manga109 datasets. Best viewed on screen.}
	\label{fig_COMSR_x4}
\end{figure*}

\noindent \textbf{Evaluation Metrics:} To quantitatively compare the recovered HR results of the proposed model with that of the state-of-the-art models, PSNR and SSIM are used, which are calculated based on the luminance channel of the YCbCr space of the recovered RGB results. 

\noindent \textbf{Training Settings:} We set the number of MRFAMs as $G=5,5,20$ in the MRFAG structure for Stage 1, Stage 2 and Stage 3, respectively. In each MRFAM, we set the number of residual blocks as $M=5$. All the convolution layers have $C=64$ filters except for those in the dilated convolution layer as shown in Fig.~\ref{fig_mrfag}, where the convolution layer has $C=4$ filters. We use $3\times3$ as the filter size for all convolution layers except for those in the dilated convolution based channel attention where the kernel sizes are $1\times1$, $3\times3$ and $5\times5$, which are shown in Fig.~\ref{fig_mrfag}. Following previous works~\cite{lim2017enhanced,zhang2018image,dai2019second}, we adopt the sub-pixel convolution~\cite{shi2016real} to upsample the LR features to HR. During training, we also augment the training dataset by randomly rotating by $90^\circ$, $180^\circ$, $270^\circ$ and flipping horizontally~\cite{lim2017enhanced,zhang2018image,dai2019second}. In each training batch, LR images with patch size $48\times48$ are cropped as inputs. The proposed model is trained by the ADAM optimizer with a fixed initial learning rate of $10^{-4}$. The whole process is implemented in the PyTorch platform with 4 Nvidia TITAN TRX GPUs, each with 24GB of memory.

\subsection{Comparisons with State-of-the-arts}
\label{sec_compar}
In this section, we compare our HIPA with 17 state-of-the-art SISR methods: SRMD~\cite{zhang2018learning} DBPN~\cite{haris2018deep}, RDN~\cite{zhang2018residual}, MSRN~\cite{li2018multi}, RCAN~\cite{zhang2018image}, SRFBN~\cite{li2019feedback}, SAN~\cite{dai2019second}, CSNLN~\cite{mei2020image}, HAN~\cite{niu2020single}, NSR~\cite{fan2020neural}, IGNN~\cite{zhou2020cross}, RFANet~\cite{liu2020residual}, NLSN~\cite{mei2021image}, SwinIR~\cite{liang2021swinir}, TDPN~\cite{cai2022tdpn}, ELAN~\cite{zhang2022efficient} and DGSM-Swin~\cite{huang2023deep}. Following previous works~\cite{zhang2018image,dai2019second,niu2020single,liang2021swinir}, we also perform self-ensemble on our HIPA to further improve its performance and dub it HIPA+.

\noindent \textbf{Quantitative Comparison:}
Table~\ref{tab_ComSR} reports the quantitative comparisons between our method and 17 state-of-the-art SISR methods on five benchmark datasets for scale factor $2\times$, $3\times$ and $4\times$. The best results are \textbf{highlighted} in \textcolor{red}{red} and the second best in \textcolor{blue}{blue}. All the reported methods are proposed in recent 5 years and have achieved competitive results. Compared with these methods, our HIPA+ achieves the best results on multiple benchmarks for all scaling factors and surpasses most state-of-the-art methods in terms of PSNR and SSIM. Without using self-ensemble our network HIPA still achieves the best results on multiple benchmarks for all scale factors. It is noteworthy that our proposed HIPA is superior to SwinIR~\cite{liang2021swinir} and DGSM-Swin~\cite{huang2023deep}, both of which are all hybrid architecture similar to HIPA. Specifically, the values of PSNR on the Urban100 dataset for scale factor $\times4$ are improved by \textbf{0.2 dB} and \textbf{0.54 dB}, respectively, compared to SwinIR and DGSM-Swin. The main reasons may lie in that i) the designed  multi-stage progressive model not only can exploit features from different size patches but also can gradually recover the HR image from coarse to fine; and ii) the proposed MRFAG can let the network exhaustively mine local features contained in the original LR image from different receptive fields based on dilated convolution with different dilation factors.

\noindent \textbf{Qualitative Comparison:} In Fig.~\ref{fig_COMSR_x4}, we also visually illustrate the zoomed in comparison results with some state-of-the-art methods on several images from the test datasets. From the results, we find that our proposed HIPA can always obtain sharper results and recover more high-frequency textures and details, while most competing SISR models suffer from some unpleasant blurring artifacts. Take ``img\_109" in Manga109 shown in Fig.~\ref{fig_COMSR_x4} as an example, existing methods obtain heavy blurring artifacts. The early proposed Bicubic fails to generate the clear structures. Although more recent methods, e.g. RCAN~\cite{zhang2018image}, SAN~\cite{dai2019second}, CSNLN~\cite{mei2020image}, RFANet~\cite{liu2020residual}, IGNN~\cite{zhou2020cross}, NSLN~\cite{mei2021image} and SwinIR~\cite{liang2021swinir} can recover the main outlines, they fail to recover textures and details, and even generate some distorted and deformed textures. In contrast, our method effectively recovers textures through using the proposed HIPA and MRFAG.
\begin{table}[!htb]
	\centering
	\scriptsize
	\renewcommand{\arraystretch}{0.9} 
	\caption{Total number of parameters, computational complexity, running time and PSNR comparison on Urban100 dataset for scale factor $\times4$ of different models.}
	\begin{tabular}{p{1.6cm}<{\centering} |p{1.04cm}<{\centering} | p{1.04cm}<{\centering} | p{1.04cm}<{\centering} | p{1.04cm}<{\centering} }
		\hline
		Model                             & Params(M) & FLOPs(G)  & Time(s)  & PSNR(dB)  \\ \hline\hline
		EDSR~\cite{lim2017enhanced}       & 43        & 2875      & 0.5437   & 26.64     \\ 
		RDN~\cite{zhang2018residual}      & 22.3      & 1305      & 0.3127   & 26.61     \\ 
		RCAN~\cite{zhang2018image}        & 16        & 912       & 0.3891   & 26.83     \\ 
		IPT~\cite{chen2021pre}            & 114       & 1480      & 1.3550   & 27.26     \\ 
		SwinIR~\cite{liang2021swinir}     & 11.8      & 978       & 0.4331   & 27.07     \\ \hdashline
		ESRT~\cite{lu2022transformer}     & 0.68      & 65.2      & 0.0106   & 26.39     \\ 
		HNCT~\cite{fang2022hybrid}        & 0.37      & 78.8      & 0.0154   & 26.20     \\ 
		Swin2SR-s~\cite{conde2023swin2sr} & 1         & 146.5     & 0.0223   & 26.58     \\\hline\hline
		HIPA(ours)                        & 11.3      & 764       & 0.2615   & 27.27     \\ \hline
	\end{tabular}
	\label{tab_ComPara}
\end{table}

\noindent \textbf{Further Comparison:}
Table~\ref{tab_ComPara} compares the number of parameters, computational complexity and average running time comparisons for various SISR methods. Except for ESRT~\cite{lu2022transformer}, HNCT~\cite{fang2022hybrid} and Swin2SR-s~\cite{conde2023swin2sr}, which are light weight SISRs, the other methods (including our HIPA) are classic performance-oriented SISRs. EDSR~\cite{lim2017enhanced}, RDN~\cite{zhang2018residual} and RCAN~\cite{zhang2018image} are CNN-based SISR methods, while IPT~\cite{chen2021pre}, SwinIR~\cite{liang2021swinir}, ESRT~\cite{lu2022transformer}, HNCT~\cite{fang2022hybrid} and Swin2SR-s~\cite{conde2023swin2sr} are state-of-the-art Transformer-based SISRs. 

Compared to EDSR~\cite{lim2017enhanced}, RDN~\cite{zhang2018residual} and RCAN~\cite{zhang2018image}, HIPA not only has fewer parameters but also achieves a better PSNR value. Compared to two similar performance-oriented methods: IPT~\cite{chen2021pre} and SwinIR~\cite{liang2021swinir}, our method is more efficient in both computational time and memory usage. Although classic performance-oriented SISRs have more parameters and high computational complexity than light weight SISRs, they have better PSNR values because they focus more on performance.  
\subsection{Ablation Study}
\begin{table}[!htb]
	\centering
	\scriptsize
	\renewcommand{\arraystretch}{0.9} 
	\caption{Ablation study of the designed multi-size patch input for the proposed method. All the experiments are conducted with the same experimental conditions except that for the patch embedding used in HIPA Transformer, and tested on the Set14 and Urban100 datasets for scale factor $\times4$.}
	\label{tab_Hier_AL}
	\begin{tabular}{p{0.12cm}<{\centering}|p{0.12cm}<{\centering}|p{0.5cm}<{\centering}|p{0.5cm}<{\centering}|p{0.5cm}<{\centering}|p{0.5cm}<{\centering}|p{0.5cm}<{\centering}|p{0.5cm}<{\centering}}
		\hline
		\multicolumn{2}{c|}{\multirow{2}{*}{Design}} & \multicolumn{3}{c|}{Set14} & \multicolumn{3}{c}{Urban 100}\\ \cline{3-8} 
		\multicolumn{2}{c|}{}                        & $\times$2 & $\times$3 & $\times$4  & $\times$2 & $\times$3 & $\times$4 \\ \hline\hline
		\multicolumn{2}{c|}{Fixed-size patch }       & 34.21     & 30.76     & 28.95      & 33.43     & 29.30     & 27.18    \\ \hline
		\multicolumn{2}{c|}{Our multi-size patch}    & 34.25     & 30.84     & 29.07      & 33.50     & 29.41     & 27.27    \\ \hline
	\end{tabular}
\end{table}

\noindent \textbf{Ablation Study of HIPA Transformer:} In Table~\ref{tab_Hier_AL}, we report the quantitative comparisons between the proposed HIPA Transformer with fixed-size patches and with multi-size patches by letting the network with and without partitioning the input LR image into a hierarchy of subblocks for scale factor $\times2$, $\times3$ and $\times4$ on the Set14 and Urban100 datasets. From the PSNR results, we find that the HIPA Transformer using patches of different sizes outperforms that using fixed-size patches by a maximum of 0.12dB. The main reason is that the hierarchy of subblocks let the network learn one LR image from different sizes and improves the overall performance of the final results. Our result not only validates the effectiveness of the proposed multi-size patch but also further validates the effectiveness of the proposed hierarchical multi-stage structure.
\begin{table}[!htb]
	\centering
	\scriptsize
	\renewcommand{\arraystretch}{0.9} 
	\caption{Impact of HIPA Transformer size for the proposed method. HIPA\_S, HIPA\_M and HIPA\_L denote small, medium and large version of HIPA Transformer, respectively. PatS, HeadN and LayerN, respectively, denote the patch size, the head number and the layer number of HIPA Transformer. All the experiments are conducted with the same experimental conditions, and tested on the Set14 dataset for scale factor $\times4$.}
	\label{tab_Model_AL}
	\begin{tabular}{p{1.5cm}<{\centering}|p{0.77cm}<{\centering}|p{0.77cm}<{\centering}|p{0.77cm}<{\centering}||p{0.77cm}<{\centering}|p{0.77cm}<{\centering}}
		\hline
		HIPA Index   & PatS  & HeadN  & LayerN  & Params & PSNR\\ \hline\hline
		HIPA\_S      & 4     & 4      & 4       & 8.9 M  & 28.99\\ \hline
		HIPA\_M      & 8     & 8      & 8       & 11.3 M & 29.07\\ \hline
		HIPA\_L      & 16    & 16    & 16       & 16.1 M & 29.12\\ \hline
		
	\end{tabular}
\end{table}

Besides, in Table~\ref{tab_Model_AL}, we show the effects of the HIPA Transformer size on model performance. It can be found that the PSNR is positively correlated with the HIPA Transformer size. Even though the performance keeps increasing, the total number of parameters of the proposed HIPA Transformer grows also. To balance the performance and model size, we choose HIPA\_M (PatS $= 8$, HeadNr $=8$ and LayerN $=8$) in the rest of the experiments. 
\begin{table}[!htb]
	\centering
	\scriptsize
	\renewcommand{\arraystretch}{0.9} 
	\caption{Ablation study on the HIPA transformer using `PE', `CPE' and the proposed `APE' for scale $\times 2$, $\times 3$ and $\times 4$ on the Manga109 dataset.}
	\label{tab_APE_AL}
	\begin{tabular}{p{1.8cm}<{\centering}|p{1.5cm}<{\centering}|p{1.5cm}<{\centering}|p{1.5cm}<{\centering}}
		\hline
		Different PE                   & $\times$2 & $\times$3 & $\times$4   \\ \hline\hline
		PE~\cite{dosovitskiy2021image} & 39.69     & 34.80     & 31.74       \\ \hline
		CPE~\cite{chu2021conditional}  & 34.71     & 34.83     & 31.77       \\ \hline
		APE (Ours)                     & 39.75     & 34.88     & 31.83        \\ \hline
	\end{tabular}
\end{table}

\noindent \textbf{Ablation Study of the Proposed APE:}  To validate the effectiveness of the proposed attention position encoding (APE), a comparison experiment between the proposed method using the previous position embedding (PE)~\cite{dosovitskiy2021image}, the condition position encoding (CPE)~\cite{chu2021conditional} and the proposed APE is conducted for scale $\times 2$, $\times 3$ and $\times 4$ on the Set14 and Urban100 datasets. From the PSNR results shown in Table~\ref{tab_APE_AL}, we find that the HIPA Transformer using the proposed APE obtains superior performance than that using the previous PE and CPE for all scales on the two datasets, which validates the effectiveness of the proposed APE.

\begin{table}[!htb]
	\centering
	\scriptsize
	\renewcommand{\arraystretch}{0.9} 
	\caption{Ablation study of the MRFAG on the B100, Urban100 and Manga109 datasets for scale factor $\times 3$.}
	\vspace{-5pt}
	\begin{tabular}{p{1.8cm}<{\centering} | p{1.5cm}<{\centering} | p{1.5cm}<{\centering} | p{1.5cm}<{\centering} }
		\hline
		Module                        & B100         & Urban100     & Manga109   \\ \hline\hline
		w/o MRFAG                     & 29.41        & 29.35        & 34.80       \\ \hline
		w/ RCAB~\cite{zhang2018image} & 29.43        & 29.37        & 34.83       \\ \hline
		w/ MRFAG                      & 29.45        & 29.41        & 34.88       \\ \hline
	\end{tabular}
	\label{tab_dilated1}
\end{table}

\noindent \textbf{Ablation Study of the Proposed MRFAG:} To validate the effectiveness of the MRFAG, as shown in Table~\ref{tab_dilated1}, we conduct a comparison experiment between the proposed method without using the proposed MRFAG, with using a classic and effective feature extraction module RCAB~\cite{zhang2018image} and with the proposed MRFAG. Note that, for a fair comparison, the total number of parameters of our method with RCAB modules is 11.6M, which is close to the number of parameters for our method with MRFAG, which is 11.3M. It is found that the improvement of the proposed method with using the proposed MRFAG is greater than that of the proposed method using the RCAB, which validates the effectiveness of the proposed MRFAG.  
\begin{table}[!htb]
	\centering
	\scriptsize
	\renewcommand{\arraystretch}{0.9} 
	\caption{PSNR comparison between the proposed method using SE attention and using the proposed convolution based attention on the B100, Urban100 and Manga109 datasets for scale factor $\times 4$.}
	\vspace{-5pt}
	\begin{tabular}{p{3cm}<{\centering} | p{1.1cm}<{\centering} | p{1.1cm}<{\centering} | p{1.1cm}<{\centering} }
		\hline
		Module                          & B100      & Urban100  & Manga109   \\ \hline\hline
		Standard SE attention           & 27.87     & 27.22     & 31.78      \\ \hline
		Dilated convolution attention   & 27.90     & 27.27     & 31.83      \\ \hline
	\end{tabular}
	\label{tab_dilated}
\end{table}

In addition, as shown in Table~\ref{tab_dilated}, another comparison between the proposed MRFAG using SE attention and using the proposed dilated convolution based attention to validate the effectiveness of the proposed dilated convolution based attention. It can be found that the proposed dilated convolution based attention can improve PSNR value by a mean 0.043 dB on the B100, Urban100 and Manga109 datasets for scale factor $\times 4$ than that of the standard SE attention.

\section{Conclusion}
\label{sec_Con}
In this paper, we propose the Hierarchical Patch Transformer (HIPA) for accurate single image super resolution, which progressively recovers the high resolution image by partitioning the input into a hierarchy of patches. Specifically, a multi-stage progressive model is employed where the earlier stages use smaller patches as tokens and the final stage operates at full resolution. Our architecture is a cascade CNNs and Transformers for feature aggregation across multiple stages. 
In addition, we develop a novel attention-based position encoding scheme that allows the Transformer focus on the important tokens and easily process an input low resolution images with varying sizes. Besides, the proposed multi-receptive field attention module can enlarge the convolution receptive field from different branches. The quantitative and qualitative evaluations on different benchmark datasets demonstrate the effectiveness of the hierarchical patch partition over using fixed-size patches, as well as the superior performance of the proposed HIPA over most state-of-the-art methods in PSNR, SSIM and visual quality.

{
	\small
	\bibliographystyle{IEEEtran}
	\bibliography{HIPA}
}
\vspace{-10 mm}
\begin{IEEEbiography}[{\includegraphics[width=1in,height=1.25in,clip,keepaspectratio]{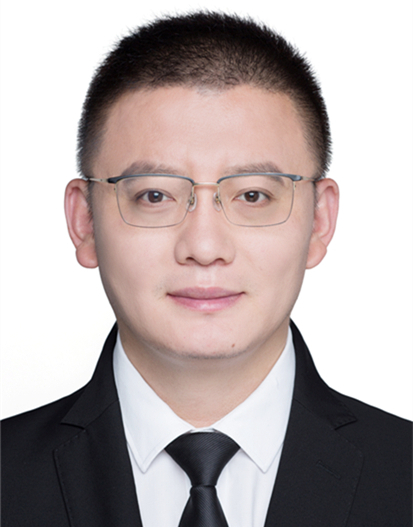}}]{Qing Cai} received the M.Sc. and Ph.D. degree from the Department of Automation, Northwestern Polytechnical University, Xi'an, China, in 2016 and 2019, respectively. From 2017 to 2018, he was a Visiting Ph.D. Student in the Department of Computing Science at University of Alberta. From 2020 to 2022, he was a Postdoctoral Fellow with The Chinese University of Hong Kong at Shenzhen, and also with the University of Science and Technology of China. He is currently an Associate Professor with the Faculty of Information Science and Engineering, at Ocean University of China. His research interests include machine learning, deep learning, and computer vision, with a focus on image restoration, image segmentation, medical image processing and visual tracking. 
\end{IEEEbiography}
\vspace{-10 mm}
\begin{IEEEbiography}[{\includegraphics[width=1in,height=1.25in,clip,keepaspectratio]{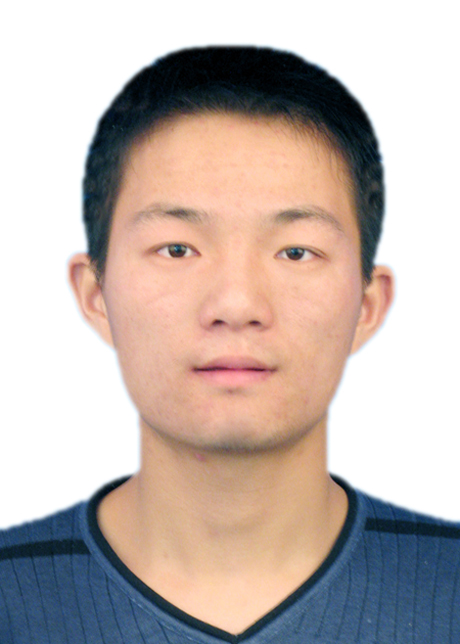}}]{Yiming Qian} received the B.Sc. degree from University of Science and Technology of China, Hefei, China, in 2012, the M.Sc. degree from Memorial University of New-foundland, St. John’s, Canada, in 2014, and the Ph.D. degree from the Department of Computing Science, University of Alberta, Edmonton, Canada, in 2019. He is currently an Assistant Professor with the University of Manitoba, Canada. His research interests include computer vision and computer graphics, while he recently focuses on 3D modeling.
\end{IEEEbiography}
\vspace{-10 mm}
\begin{IEEEbiography}[{\includegraphics[width=1in,height=1.25in,clip,keepaspectratio]{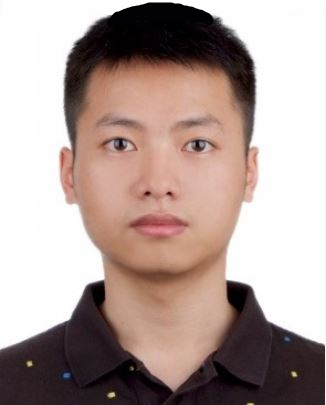}}]{Jinxing Li}  received the B.Sc. degree from the Department of Automation, Hangzhou Dianzi University, Hangzhou, China, in 2012, the M.Sc. degree from the Department of Automation, Chongqing University, Chongqing, China, in 2015, and the Ph.D. degree from the Department of Computing, Hong Kong Polytechnic University, Hong Kong, in 2018. He is currently an Associate Professor with the Harbin Institute of Technology at Shenzhen. His research interests are pattern recognition, deep learning, medical biometrics, and machine learning.
\end{IEEEbiography}
\vspace{-10 mm}
\begin{IEEEbiography}[{\includegraphics[width=1in,height=1.25in,clip,keepaspectratio]{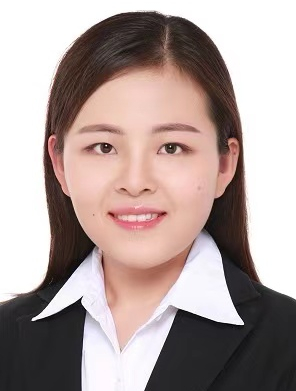}}]{Jun Lyu}  received the B.Sc. degree in intelligence science and technology from Xidian University, in 2013, and the Ph.D. degree in biomechanics and medical engineering from Peking University, in 2018. From October 2017 to April 2018, she was a Visiting Ph.D. Student in the David Geffen School of Medicine at University of California, Los Angeles. She is currently a Postdoctoral Fellow with the School of Nursing, The Hong Kong Polytechnic University. Her research interests include deep learning, and medical image processing and analysis.
\end{IEEEbiography}
\vspace{-10 mm}
\begin{IEEEbiography}[{\includegraphics[width=1in,height=1.25in,clip,keepaspectratio]{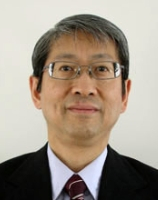}}]{Yee-Hong Yang}(Senior Member, IEEE) received the B.Sc. (first honors) from the University of Hong Kong, the M.Sc. from Simon Fraser University, and the Ph.D. from the University of Pittsburgh. He was a faculty member in the Department of Computer Science at the University of Saskatchewan from 1983 to 2001 and served as Graduate Chair from 1999 to 2001. While there, in addition to department level committees, he also served on many college and university level committees. Since July 2001, he has been a Professor in the Department of Computing Science at the University of Alberta. He served as Associate Chair (Graduate Studies) in the same department from 2003 to 2005. His research interests cover a wide range of topics from computer graphics to computer vision, which include physically based animation of Newtonian and non-Newtonian fluids, texture analysis and synthesis, human body motion analysis and synthesis, computational photography, stereo and multiple view computer vision, and underwater imaging. He has published over 100 papers in international journals and conference proceedings in the areas of computer vision and graphics. He is a Senior Member of the IEEE and serves on the Editorial Board of the journal Pattern Recognition. In addition to serving as a reviewer to numerous international journals, conferences, and funding agencies, he has served on the program committees of many national and international conferences. In 2007, he was invited to serve on the expert review panel to evaluate computer science research in Finland.
\end{IEEEbiography}
\vspace{-10 mm}
\begin{IEEEbiography}[{\includegraphics[width=1in,height=1.25in,clip,keepaspectratio]{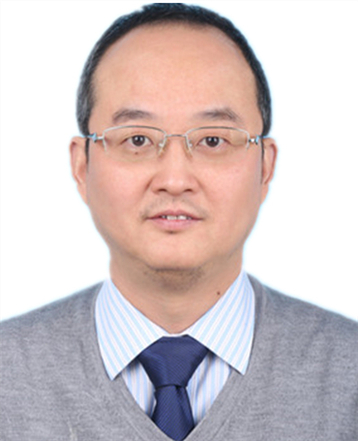}}]{Feng Wu}(Fellow, IEEE) received the B.Sc. degree in electrical engineering from Xidian University, Xi’an, China, in 1992, and the M.Sc. and Ph.D.degrees in computer science from the Harbin Institute of Technology, Harbin, China, in 1996 and 1999, respectively. He was a Principle Researcher and a Research Manager with Microsoft Research Asia, Beijing, China. He is currently a Professor with the University of Science and Technology of
	China, Hefei, China, where he is also the Dean of the School of Information Science and Technology. His research interests include image and video compression, media communication, and media analysis and synthesis.
\end{IEEEbiography}
\vspace{-10 mm}
\begin{IEEEbiography}[{\includegraphics[width=1in,height=1.25in,clip,keepaspectratio]{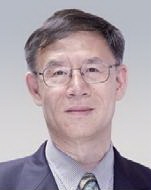}}]{David Zhang}(Life Fellow, IEEE) graduated in Computer Science from Peking University. He received his MSc in 1982 and his PhD in 1985 in both Computer Science from the Harbin Institute of Technology (HIT), respectively. From 1986 to 1988 he was a Postdoctoral Fellow at Tsinghua University and then an Associate Professor at the Academia Sinica, Beijing. In 1994 he received his second PhD in Electrical and Computer Engineering from the University of Waterloo, Ontario, Canada. He has been a Chair Professor at the Hong Kong Polytechnic University where he is the Founding Director of Biometrics Research Centre (UGC/CRC) supported by the Hong Kong SAR Government since 1998. Currently he is Presidential Chair Professor in Chinese University of Hong Kong (Shenzhen). So far, he has published over 20 monographs, 500+ international journal papers and 40+ patents from USA/Japan/HK/China. He has been continuously 8 years listed as a Global Highly Cited Researchers in Engineering by Clarivate Analytics during 2014-2021. He is also ranked about 85 with H-Index 123 at Top 1000 Scientists for international Computer Science and Electronics. Professor Zhang is a Croucher Senior Research Fellow, Distinguished Speaker of the IEEE Computer Society, and Fellows of both Royal Societ of Canada and Canadian Academy of Engineering, as well as IEEE Life Fellow and IAPR/AAIA Fellow.
\end{IEEEbiography}

\end{document}